\documentclass[conference]{IEEEtran}

\usepackage{graphicx}
\usepackage{booktabs}
\usepackage{amsmath,amsthm,amssymb}
\usepackage{dsfont}
\usepackage{bbold}

\usepackage{multirow}

\newtheorem{definition}{Definition}

\newtheorem{example}{Example}

\newcommand{\comment}[1]{}

\newcommand{\entity}[1]{\textsf{#1}}

\begin{document}

\title{From Mastery Profile to Simulated Response: Stochastic Student Knowledge 
Graphs (SSKG) for Faithful LLM Student Simulation}

\author{\IEEEauthorblockN{Yuan An, Emily Wang\textsuperscript{\textsection},
Benjamin Wang\textsuperscript{\textsection},
Ruhma Hashmi}
\IEEEauthorblockA{\textit{School of Computer and Information Sciences, Nick Howley College of Engineering and Computing} \\
\textit{Drexel University}\\
Philadelphia, PA 19104, USA \\
\{ya45, rh927\}@drexel.edu}}

\maketitle

\begingroup\renewcommand\thefootnote{\textsection}
\footnotetext{Drexel summer interns.}
\endgroup

\begin{abstract}
Large language models (LLMs) are increasingly used to simulate students at 
different mastery levels to generate synthetic training data and 
stress-test tutoring systems. Existing approaches primarily leverage an 
LLM to answer questions or perform tests while instructing it to simulate various skill levels.
However, LLMs tend to execute at their full built-in capability regardless of 
target skill levels. We demonstrate this limitation on SAT questions 
across five mastery levels, where three LLMs consistently achieve 96.8–100\% accuracy.
To faithfully simulate student behavior, we propose a Stochastic Student 
Knowledge Graph (SSKG)-based approach that leverages the strengths of LLM in different ways. 
Using human-in-the-loop workflows, 
LLMs extract a Curriculum Knowledge Graph (CKG) from course materials, 
decompose solution paths into CKG triples, and map incorrect distractors. 
A student’s mastery profile is represented as an SSKG, assigning mastery 
probabilities to CKG triples. When solving a problem, the SSKG is 
sampled to determine the outcome, after which an LLM generates 
a rationale consistent with the sampled result.
Compared to seven baselines which fail to separate profiles and maintain 86.5–100\% 
accuracy, our SSKG-based approach achieves 44.1–85.2\% accuracy from 
low to high-mastery profiles, creating a clear monotonic gradient. 
An ablation study demonstrates that taxonomy awareness 
contributes most significantly to these fidelity gains, 
particularly for lower-mastery profiles.

\end{abstract}


%
%
\section{Introduction}
\label{sec:intro}

\begin{figure}[t]
	\centering
  	\includegraphics[width=\columnwidth]{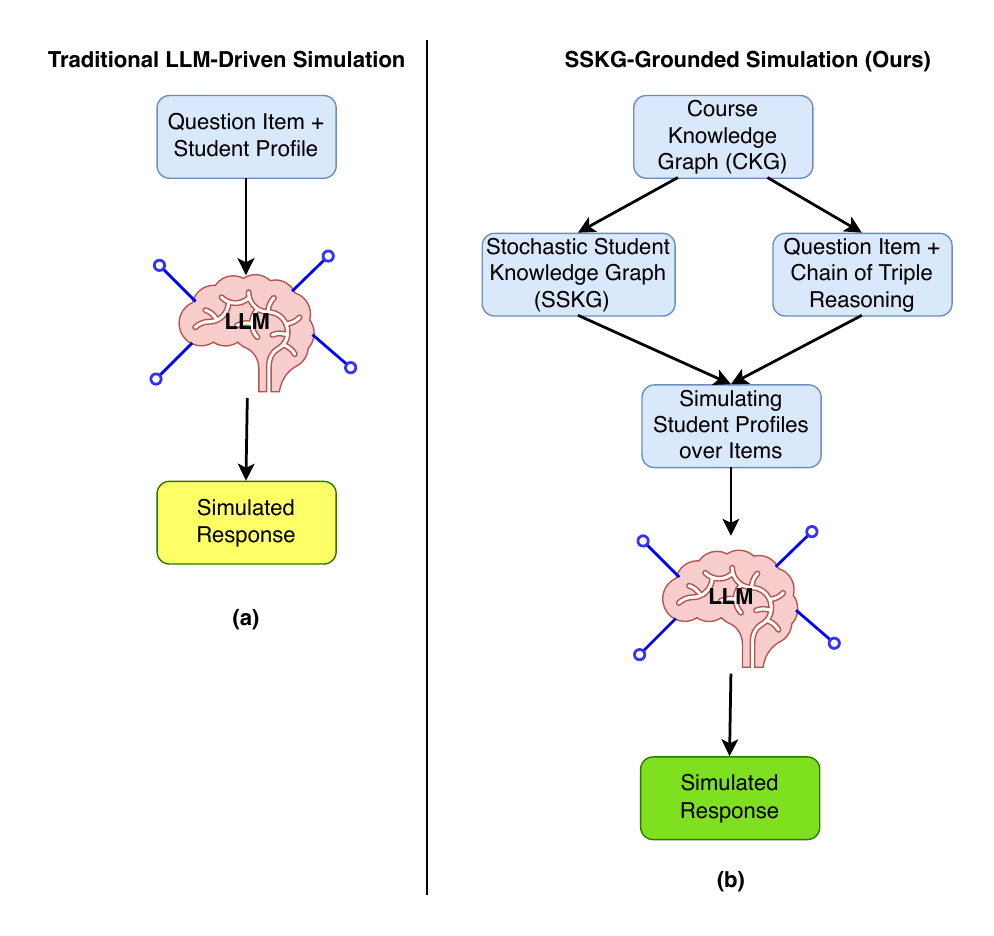}
	\caption{Traditional LLM-driven student simulation (a) versus SSKG-grounded simulation (b). In the traditional pipeline (a), 
	a question item and a student profile are handed directly to an LLM, which decides both the simulated 
	answer and its correctness. In our pipeline (b), a Curriculum Knowledge Graph (CKG) is built as a set of 
	fine-grained triples extracted from the source curriculum. A student profile is represented as a Stochastic 
	Student Knowledge Graph (SSKG), which assigns a mastery probability to each CKG triple at a time point. Each question is 
	mapped to the chain of triples required to solve it. We simulate a student profile over an item by sampling 
	over the item's chain against the SSKG's mastery probabilities to decide whether the response is correct. The resulting 
	item paired with this correctness label is then passed to the LLM, which narrates a first-person rationale consistent 
	with the outcome.}
	\label{fig:overview}
\end{figure}

Synthetic student response data is valuable when real classroom data is expensive, slow, or 
difficult to collect at scale due to various constraints. 
Applications using synthetic student data include training knowledge-tracing models before a course 
has run \cite{jung2025clst}, stress-testing an intelligent tutoring
system across a spread of mastery levels \cite{noh2026simulating}, or piloting new 
assessment items before they reach real students \cite{sauberli2025psychometrically}.
LLMs are an attractive tool for generating such data because they can answer 
a wide range of questions and explain their reasoning in natural 
language \cite{openai2023gpt4}. 
The open question is: \emph{ 
how to leverage an LLM that can solve a curriculum’s 
questions essentially perfectly \cite{dewinter2024system2} 
to simulate students at different mastery levels?}

\noindent
\textbf{Example.} \emph{Given a set of SAT Algebra questions, we would like to simulate three types of students
who have mastered the algebra knowledge required by these questions at a high, a medium, and a low
level. A faithful simulation should then answer a high percentage of the questions correctly for the
first student, a medium percentage for the second, and a low percentage for the third. Moreover, 
if a student has a gap in a specific part of the curriculum, their simulated errors should concentrate on 
questions that require that knowledge. Each incorrect answer should also be traceable to the specific 
algebra concept or skill the student failed to apply.} $\square$

The common LLM-based approach in the literature and in practice is to describe 
the desired student in the prompt, for example, by specifying a skill level~\cite{benedetto2024using},
an exam grade~\cite{srivatsa2025can}, or an archetypal student profile~\cite{luwang2024generative}, and let the LLM
decide both the answer and the explanation (Figure~\ref{fig:overview} (a)).
We show this approach fails on a concrete, realistic benchmark. 
Using 379 SAT Algebra multiple-choice questions from the College Board question 
bank \cite{collegeboard2026sat} 
and five archetypal profiles ranging from near-expert to severely gapped, 
three LLMs spanning three vendors answer 
at 96.8-100\% accuracy \emph{regardless of profile}, including a 
``struggling student who has 
low mastery of most algebra concepts.'' 
This is not a new observation in isolation, it replicates and 
extends a competency-bias finding first reported for GPT-3.5 on non-math 
domains~\cite{benedetto2024using}.

To address the problem, we introduce a simulation method that is grounded in 
the concept of Stochastic Student Knowledge Graph (SSKG) 
(see Figure~\ref{fig:overview} (b)).
After constructing a curriculum knowledge graph (CKG) containing a set of fine-grained triples, we represent a student profile as
an SSKG which assigns a mastery probability to each CKG triple at a time point. Furthermore, we characterize a question's solution as a chain of 
triples identified from the CKG. To simulate, we check the required chain of triples of a question against the profile's mastery vector by 
explicit sampling. This analytic step decides both whether the simulated student answers correctly and, on failure, 
which specific knowledge unit broke first. Finally, an LLM is invoked to render a first-person rationale consistent with 
the already-decided outcome. 

We design this SSKG-based method into four cumulative ablation arms (a-d) so that any gain can be attributed to
a specific mechanism. 
\texttt{arm-a} is a single-draw sampler. One Bernoulli draw is made per required triple,
so every triple in an item's chain is exactly as consequential as every other. 
\texttt{arm-b} adds a retrieval/execution (knowledge-tracing) decomposition, separating whether
a fact is available to the student at all (retrieval) from whether it is applied without error
(execution). 
\texttt{arm-c} adds taxonomy-conditioned item weighting so that failing a purely 
definitional fact (e.g., forgetting what a variable is) is not as consequential
as failing a procedural or relational one (e.g., misapplying a multi-step method or picking the wrong solution
approach).
\texttt{arm-d} adds distractor routing. Rather than selecting
uniformly a wrong choice, a failure at a specific triple selects the
specific wrong answer choice that failure would produce using an item's distractor map.

We make the following contributions. 
\begin{enumerate}

\item We develop an SSKG-based method to simulate students with stratified profiles. The method makes
profile-accuracy monotonicity and skill-specific error patterns.
\item We present both the theoretical foundation and practical implementation of the simulation method.
\item We compare the method against seven published and standard baselines and a
scatter control on the same items, profiles, and LLM. We use four metrics, M1 to M4, each tied
to one property of faithful simulation. We report where SSKG separates clearly from every
baseline.
\item We conduct a four-arm cumulative ablation study that attributes the method's gains to
specific mechanisms.

\end{enumerate}

The rest of the paper is organized as follows.
Section~\ref{sec:related} discusses the related work.
Section~\ref{sec:example} introduces a running illustrative SAT example.
Section~\ref{sec:formalism} formalizes several key concepts.
Section~\ref{sec:problem} defines the problem of faithful simulation.
Section~\ref{sec:object-construction} constructs the elements used in the simulation. 
Section~\ref{sec:simulation} presents the theoretical foundations and implementation.
Section~\ref{sec:evaluation} compares the method against the baselines and reports the ablation study.
Section~\ref{sec:discussion} discusses the findings.
Section~\ref{sec:limitations} describes the limitations.
Finally, Section~\ref{sec:conclusion} concludes the paper and discusses future directions.


%
%
\section{Related Work}
\label{sec:related}


\subsection{Educational Knowledge Graphs} 

MOOCCube~\cite{yu2020mooccube} builds a large-scale 
knowledge graph of course concepts, videos, and exercises from over 700 real MOOCs specifically to support 
downstream NLP and learning-analytics tasks.
Pan et al.~\cite{pan2024-unifying} catalog LLM-for-KG-construction patterns at length.
Because an LLM-proposed triple is not automatically a correct one, a human verification is needed. 
Tsaneva et al.~\cite{tsaneva2025knowledge} show that standalone LLM validation is weak, 
while hybrid human–LLM validation provides the best overall results.

\subsection{Stochastic and Probabilistic Knowledge Graphs} 

Attaching a probability to a fact rather than treating it as universally true has been studied at several levels of 
representation. Probabilistic ontology languages such as PR-OWL~\cite{costa2006prowl} 
extend OWL (Web Ontology Language) with 
Bayesian semantics so that class membership and relations can be 
asserted with a degree of belief rather than as certain axioms. 
Probabilistic knowledge graphs built by open information extraction, such as 
NELL~\cite{carlson2010toward}, attach a confidence score to each automatically-extracted belief as 
it is read from text and promote only beliefs whose confidence clears a threshold. 
Uncertain knowledge graph embedding~\cite{chen2019embedding} attaches a confidence score to each 
triple and learns embeddings that predict this confidence for unseen triples. 
At a general level, a stochastic process~\cite{karlin1975first} is a collection of 
random variables indexed by a parameter. Here, the parameter is the step 
in the chain being resolved, and our per-triple sampling provides a concrete 
instance of this process.

\subsection{Knowledge Tracing and Cognitive Diagnosis} 

Modeling what a student does and does not know is a decades-old problem independent of LLMs, 
originally motivated by intelligent tutoring systems (ITS) that adapt instruction to an inferred mastery state. 
VanLehn~\cite{vanlehn2011relative} ran a meta-analysis of tutoring studies and found that step-based ITS tutoring already 
approaches the effectiveness of human tutors.
Two families of models dominate how that mastery state is inferred. Corbett and Anderson~\cite{corbett1995knowledge} 
introduced Bayesian Knowledge Tracing (BKT). Piech et al.~\cite{piech2015deep} instead introduced Deep Knowledge 
Tracing (DKT), replacing BKT's per-skill Bayesian update with a recurrent network trained end-to-end on interaction sequences.
On the cognitive-diagnosis side, Junker and Sijtsma~\cite{junker2001cognitive} formalized 
the Deterministic-Input, 
Noisy-``And''-gate (DINA) model over a binary attribute (here, triple) mastery vector. 
De la Torre~\cite{delatorre2011gdina} generalized DINA into G-DINA (Generalized DINA). Underlying 
both families is classical Item 
Response Theory (IRT)~\cite{lord1980applications}, which models a student's probability 
of a correct response as a function of 
a latent ability parameter and item-level difficulty/discrimination parameters.
Harwell et al.~\cite{harwell1996monte} showed stochastically sampling a 
simulated examinee's response to a test 
item. Their Monte Carlo IRT studies draw a latent ability for 
each of many simulated examinees and stochastically generate each 
one's response to every item from the fitted item 
parameters.

\subsection{LLM-Driven Student Simulation}

Prior LLM student simulation methods control the simulated mastery level at one of 
following loci: 
the prompt, the model's weights or logits, or an external symbolic architecture.

\textbf{Prompt-level control.} Benedetto et al.~\cite{benedetto2024using} prompt GPT-3.5 with a numeric 
skill level $\ell\in\{1,\dots,5\}$ and ask it to pick the answer choice that student would give.
They report that the unconstrained model and their highest simulated level differ by only 0.92 versus 0.90 accuracy, i.e., 
prompting compresses behavior toward the ceiling, and that the prompt is fragile to inconsequential rewording. 
Lu and Wang~\cite{luwang2024generative} instead describe the profile as a list of knowledge components (KCs) 
the student has mastered, is confused about, or has no evidence on. They ask an LLM to 
\emph{predict} a student's answer rather than role-play the student directly.  Evaluated 
against 100 real students on 20 items, the frame showed profile-aligned predictions reach 85.2\% accuracy on 
mastered KCs versus 11.0\% on confused ones. 
Acquaye et al.~\cite{acquaye2026calculators} estimate item difficulty by aggregating LLM-simulated 
response accuracy across many simulated test-takers and comparing the result to IRT-calibrated ground truth.
Srivatsa et al.~\cite{srivatsa2025can} provide independent, real-world evidence that LLM-based 
student simulations fail to preserve differences in student mastery levels.

\textbf{Model-level control.} PS$^2$~\cite{liu2026ps2} interpolates, at the logits level, between a
strong upper-bound LLM and a weak lower-bound LLM fine-tuned to commit conceptual and procedural
errors. They demonstrate that this preserves monotone accuracy ordering at a finer granularity than prompt baselines, 
which invert orderings at five to seven levels. 

\textbf{Architectural control.} BEAGLE~\cite{wang2026beagle} is the closest precedent to our 
design commitment that architecture, not prompting, should enforce fidelity. It applies 
a semi-Markov controller over metacognitive states, Bayesian Knowledge Tracing (BKT) with explicit flaw 
injection, and a decoupled strategist/executor. 
On open-ended Python problem solving, BEAGLE's epistemic-fidelity error-recurrence rate (86.2\%) 
sits well above nine prompting baselines (as low as 7.8\% for vanilla prompting). 

%
%
\section{Illustrative Example}
\label{sec:example}

Table~\ref{tab:example} presents an SAT Algebra item, 
including question stem, four answer choices, 
correct answer, College-Board-assigned 
skill and difficulty, and its official rationale.

Two features make this item a useful running example. First, the 100 miles/week figure in the stem is itself 
a distractor at the reading-comprehension level. It plays no role in the correct equation. 
Second, its three wrong choices are produced by two distinct, independently identifiable single-step 
errors, an inverted unit rate (A, B) and a mistranslated right-hand side (A, C). 
We can trace exactly which curriculum fact corresponds to a simulated failure.

\begin{table}[!h]
\centering
\caption{Illustrative example: An SAT Algebra Question.}
\label{tab:example}
\footnotesize
\begin{tabular}{@{}p{1.5cm}p{6.1cm}@{}}
\toprule
\textbf{Question Stem} & \emph{Alan drives an average of 100 miles each week. His car can travel an 
average of 25 miles per gallon of gasoline. Alan would like to reduce his weekly expenditure on gasoline 
by \$5. Assuming gasoline costs \$4 per gallon, which equation can Alan use to determine how many 
fewer average miles, $m$, he should drive each week?} \\
\midrule
\textbf{Choices} & (A) $\tfrac{25}{4}m=95$ \quad (B) $\tfrac{25}{4}m=5$ \newline (C) $\tfrac{4}{25}m=95$ \quad (D) $\tfrac{4}{25}m=5$ \\
\midrule
\textbf{Correct answer} & D \\
\midrule
\textbf{Skill} & Linear equations in one variable \\
\midrule
\textbf{Difficulty} & Hard (College-Board-calibrated) \\
\midrule
\textbf{Rationale} & \emph{Choice D is correct. Since gasoline costs \$4 per gallon, and since 
Alan's car travels an average of 25 miles per gallon, the expression $4/25$ gives the cost, in 
dollars per mile, to drive the car. Multiplying by $m$ gives the cost for Alan to drive $m$ miles 
in his car. Alan wants to reduce his weekly spending by \$5, so setting $4/25\,m$ equal to 5 gives 
the number of miles, $m$, by which he must reduce his driving.} \\
\bottomrule
\end{tabular}
\end{table}

%
%
\section{Core Formalisms of Faithful Simulator}
\label{sec:formalism}

To faithfully predict a student's response to a curriculum item, 
a simulator must evaluate the item against the student's current 
mastery level. This process requires four essential inputs: the overall 
body of curriculum knowledge, the specific knowledge chain needed to 
answer the item, a map connecting knowledge-level misconceptions 
to plausible wrong answers, and the student's mastery state. 
We formalize these elements as Curriculum Knowledge Graph (CKG), Knowledge Graph (KG) Triple Chain, Distractor Map, and Stochastic Student Knowledge Graph (SSKG).

\begin{definition}[Curriculum Knowledge Graph (CKG)]
A CKG is a pair $(T, \mathrm{pos})$ where $T=\{t_1,\dots,t_n\}$ is a set of triples 
$t_i=(\mathrm{subj}_i,\mathrm{pred}_i,\mathrm{obj}_i)$ and 
$\mathrm{pos}: T \to \mathbb{N}\times\mathbb{R}$ maps each triple to a 
$(\mathrm{chapter}, \mathrm{section})$ position. 
This induces a curriculum-order preorder $\preceq$ on $T$ by lexicographic comparison,

\begin{equation}
t_i \preceq t_j \iff \mathrm{pos}(t_i) \le_{\mathrm{lex}} \mathrm{pos}(t_j),
\label{eq:preorder}
\end{equation}

with $t_i \sim t_j$ (an unordered layer) when $\mathrm{pos}(t_i)=\mathrm{pos}(t_j)$. 
\label{def:ckg}
\end{definition}

\begin{figure}[!h]
\centering
\includegraphics[width=\columnwidth]{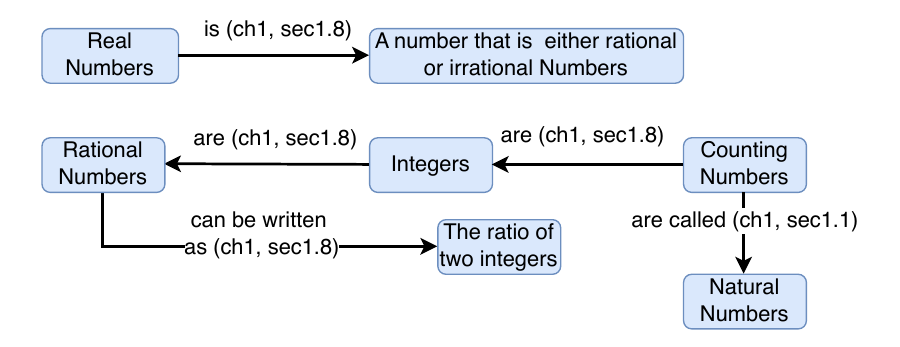}
\caption{An excerpt of a CKG's number-system connections}
\label{fig:ckg-concept}
\end{figure}

\begin{example} [CKG]
\label{ex:CKG}
Figure \ref{fig:ckg-concept} shows an excerpt of the CKG extracted from an
algebra textbook. Each triple is annotated with its $(\mathrm{chapter}, \mathrm{section})$ 
position, for example, $(\mathrm{ch1}, \mathrm{sec1.8})$.
\end{example}

\begin{definition}[KG Triple chain]
For an item $q$, the chain $\mathrm{chain}(q) = (t_1,\dots,t_k)$ is an ordered sequence of distinct triples drawn 
from $T$ in Definition~\ref{def:ckg}. Since $T$ may miss some triples in solving $q$, we 
denote $\mathrm{gap}(q)$ as the set of steps required by $q$ but not backed by any triple in $T$. 
Together, $\mathrm{chain}(q)$ and $\mathrm{gap}(q)$ account for every step of $q$'s solution derivation.
\end{definition}

\begin{example} [KG Triple Chain]
\label{ex:triple-chain}
Table \ref{tab:worked-triples} contains the triple chain of the
illustrative item in Section \ref{sec:example}. The triples are
identified from the CKG as introduced in Example \ref{ex:CKG}. 
Each triple has a unique id in the CKG and its predicate is classified into
different types (details in Section \ref{subsec:ckg}.)
\end{example}

\begin{table}[!h]
\centering
\caption{Excerpt of the illustrative item's triple chain.}
\label{tab:worked-triples}
\footnotesize
\begin{tabular}{lp{5.3cm}c}
\toprule
Id & Abbreviated triple (\textsf{subject}, \texttt{predicate}, \textsf{object}) & Type \\
\midrule
T19 & (\textsf{Variable},  \texttt{is},  \textsf{A letter representing a changeable number}) & def. \\
T149 & (\entity{Converting units}, \texttt{uses},  \entity{The identity property of mult.}) & rel. \\
T150 & (\entity{Making unit conversions}, \texttt{has\_step\_1}, \entity{Multiply by a unit fraction}) & proc. \\
T170 & (\entity{Translating a sentence}, \texttt{has\_step\_1}, \entity{Locate the ``equals'' word}) & proc. \\
T171 & (\entity{Translating a sentence}, \texttt{has\_step\_2}, \entity{Translate the left side}) & proc. \\
T172 & (\entity{Translating a sentence}, \texttt{has\_step\_3}, \entity{Translate the right side}) & proc. \\
\bottomrule
\end{tabular}
\end{table}

\begin{definition}[Distractor Map]
For an item $q$ with answer choices $\mathrm{choices}(q)$ and correct choice $y^*(q)\in\mathrm{choices}(q)$,
a distractor map is a partial function
\begin{equation}
\mathrm{dmap}(q): \mathrm{chain}(q) \rightharpoonup \mathrm{choices}(q)\setminus\{y^*(q)\},
\label{eq:dmap}
\end{equation}
assigning to each mapped triple $t_i\in\mathrm{chain}(q)$ the specific wrong choice produced when
$t_i$'s step, and only that step, fails. A triple outside the domain of $\mathrm{dmap}(q)$ is unmapped;
a failure localized there falls back to a uniform draw over $\mathrm{choices}(q)\setminus\{y^*(q)\}$.
\end{definition}

\begin{figure}[!h]
\centering
\includegraphics[width=\columnwidth]{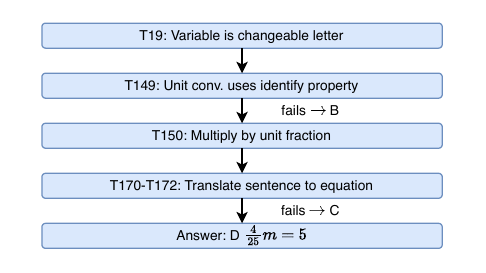}
\caption{The illustrative item's triple chain and distractor map: failures localized at T149/T150 and
T170--T172 route to choices B and C, respectively.}
\label{fig:chain}
\end{figure}

\begin{example} [Distractor Map]
\label{ex:distractor-map}
Figure~\ref{fig:chain} traces the illustrative item's chain against its distractor map: a failure at
T149/T150 (the unit-conversion step) routes to choice B, and a failure at T170--T172 (the sentence-to-equation
translation) routes to choice C. T19 is unmapped, since failing at this definitional step alone does not
reproduce any distractor's exact value. Choice A is likewise outside $\mathrm{dmap}(q)$'s range, since it
compounds both errors simultaneously and so cannot be attributed to a single triple's failure.
\end{example}

\begin{definition}[Stochastic Student Knowledge Graph (SSKG)]
\label{def:sskg}
A stochastic student knowledge graph (SSKG) represents a student's profile as mastery
probabilities over the CKG's triples: at any time point $\theta$, each triple $t\in T$ carries a
mastery probability $p(t,\theta)\in[0,1]$, and the vector $\big(p(t,\theta)\big)_{t\in T}$ changes
from one time point to the next as the student gains or forgets a piece of knowledge. This study
considers the single time point at which a student answers an SAT item; fixing $\theta$ and
suppressing it from the notation, a profile $\pi$ at simulation time is a pair
$(p_\pi, p_{\mathrm{gap},\pi})$ where $p_\pi: T\to[0,1]$ assigns each triple a mastery
probability and $p_{\mathrm{gap},\pi}\in[0,1]$ is a single fixed execution rate applied to every
known-gap step under $\pi$. 
\end{definition}

\begin{example} [SSKG]
The curriculum knowledge graph (CKG) for algebra contains 864 triples.
A student's profile at the moment of answering is then an SSKG assigning a mastery probability to
each of these 864 triples, that is, an 864-dimensional vector $p_\pi$. For instance, an average
student is represented by drawing each of the 864 values from a uniform distribution $U(0.50, 0.70)$.
\end{example}

%
%
\section{The Problem of Faithful Simulation}
\label{sec:problem}

Given a profile $\pi$ and an item $q$, let $\mathrm{correct}(q,\pi) := \mathds{1}[y(q,\pi) = y^*(q)]$ denote
whether the produced response $y(q,\pi)$ matches item $q$'s correct choice $y^*(q)$. The task is
to produce a predicted response $y(q,\pi) \in \mathrm{choices}(q)$ and a rationale $r(q,\pi)$,
satisfying the following properties:
\begin{IEEEdescription}[\IEEEsetlabelwidth{(Prop1)}]
\item[(\textbf{Prop1})] \emph{Monotonicity.} For profiles $\pi,\pi'$ with $p_\pi(t)\ge p_{\pi'}(t)$ for all $t\in T$,
$\mathbb{E}[\mathrm{correct}(q,\pi)]\ge\mathbb{E}[\mathrm{correct}(q,\pi')]$ for every $q$.
\item[(\textbf{Prop2})] \emph{Skill specificity.} For a profile whose mastery vector partitions $T$ into a
strong region and a weak region, per-item accuracy differs systematically between items whose
chain draws mostly on the strong region versus the weak region.
\item[(\textbf{Prop3})] \emph{Diagnosability.} An incorrect $y(q,\pi)$ can be attributed to a specific $t^*\in\mathrm{chain}(q)$
whose low mastery under $\pi$ produced the failure.
\end{IEEEdescription}

Before developing the SSKG-based method to solves the
\textbf{Problem of Faithful Simulation}, we first present the construction 
of the elements used by the method through human-in-the-loop workflows.

%
%
\section{Construction of the Elements used by the Faithful Simulator}
\label{sec:object-construction}

\subsection{Constructing CKG for SAT Algebra Items}
\label{subsec:ckg}

For assessing SAT algebra items, we construct the curriculum knowledge graph (CKG) 
from an open algebra textbook through a chapter-to-section-to-triple pipeline.
The text was first split into chapters and then sections, 
each section was cleaned to retain only its expository content. Then, we
extract $\langle$subject, predicate, object$\rangle$ triples using an LLM-assisted, human-verified 
process seeded by a manually-extracted convention set. 
Finally, the resulting CKG was 
checked to achieve the right rate-distortion balance, retaining the right amount of information 
from the source \cite{an2025rate}. The resulting CKG contains 864 triples spanning ten 
chapters, with each triple tagged by the chapter and section it came from.

Each triple's predicate is further classified into one of three types by a taxonomy 
built over the 163 distinct predicates in $T$: \emph{definitional} (e.g. \texttt{is}, \texttt{is\_called}, 
vocabulary that even weak students usually retain, 461 triples), \emph{procedural} 
(e.g. \texttt{has\_step\_$N$}, executable steps whose failure changes a computed 
result, 302 triples), and \emph{relational} (e.g. \texttt{uses}, \texttt{needs}, method-selection 
facts whose failure sends the student down the wrong approach entirely, 101 triples). 

\begin{figure}[t]
\centering
\includegraphics[width=\columnwidth]{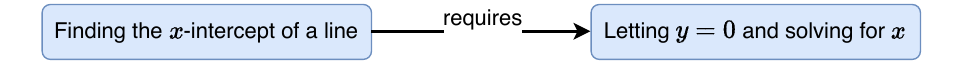}
\caption{A procedure node: a precondition triple for the \entity{Finding the $x$-intercept of a line} procedure.}
\label{fig:ckg-procedure}
\end{figure}

\textbf{Concept and Procedure nodes:}
There are two types of nodes in the CKG. Some triples' subjects/objects denote persisting mathematical objects or categories, 
for example,  \textsf{Real Numbers}, \textsf{Rational Numbers}, \textsf{Integers} in Figure~\ref{fig:ckg-concept}.
These nodes are connected through \emph{definitional} predicates like (\texttt{is}, \texttt{are}, \texttt{can\_be\_written\_as}). 
Other triples' subjects/objects denote a procedure itself, expressed as a gerund or infinitive phrase (e.g.
\entity{evaluating an expression}). 
Figure~\ref{fig:ckg-procedure} shows a single such triple, 
a precondition fact rather than a persisting object. 
We treat this as a property 
of the predicate rather than a separate node-type 
field in the CKG's schema. Procedure-denoting concepts are
overwhelmingly paired with \emph{prodecural} predicates like \texttt{has\_step\_$N$}.

%
%
\subsection{Identifying KG Triple Chain}
\label{subsec:chains}

For each assessment item, we identified a chain of required CKG triples through the human-AI
collaboration workflow described below.

\subsubsection{AI Annotation}

For a given item, an annotator (LLM-assisted) jointly reads the question stem, the answer
choices, and the item's official rationale, then walks through the rationale's derivation
step by step, matching each step against the CKG to locate the triple whose subject,
predicate, and object support it. Applying this chain of triples in sequence derives the
item's correct answer.

For example, Table~\ref{tab:worked-triples} lists the triples identified for the
illustrative item. A few representative steps show the matching process: the rationale's
``$4/25$ gives the cost... in dollars per mile'' step is supported by T149/T150 (unit
conversion via the identity property of multiplication), while its implicit ``set up the
equation from the English sentence'' step is supported by T170--T172 (the three steps of
translating a sentence into an equation).

The illustrative item also contains one \emph{known-gap} step, which composes a \$/gallon
rate and a mi/gallon rate into a combined \$/mile rate; the CKG approximates this composition
through its unit-conversion triples but does not state it as a general principle.

\subsubsection{Human verification}

Every candidate chain is then human-reviewed against three criteria: (i) 
each triple genuinely justifies the derivation step it is attached to; 
(ii) the chain is complete, every step in the official rationale is 
either covered by a triple or a known gap; and (iii) 
no triple is included that the derivation does not actually depend on. 

%
%
\subsection{Constructing the Distractor Map}
\label{subsec:distractor-map}

To identify a curriculum triple to the specific wrong-answer choice 
its failure would produce, we build such a distractor map 
also through human-AI collaboration.

\subsubsection{Annotation}
For each item, given the correct triple chain, an LLM-assisted annotator reverse-engineers
every wrong choice from two directions: forward from the item's official rationale, which often
explains why a specific distractor is wrong and therefore what mistake it encodes, and backward
from the distractor's value itself, comparing it against the correct answer to hypothesize which step
was misapplied (an inverted fraction points to a unit-conversion step, a sign flip to a negation
step, and so on). For the illustrative item, the following shows some example steps landing on distractors:
choice B ($25/4\,m=5$) inverts the unit rate at
T149/T150, and choice C ($4/25\,m=95$)
mistranslates the right-hand side at T170--T172; choice A
compounds both errors simultaneously and so is not attributable to any single triple.

\subsubsection{Verification}
Every proposed distractor map is the verified by a human reviewer to ensure that
executing the chain with deliberately wrong steps produces the wrong choice.

%
%

\subsection{Representing Profiles as SSKGs}
\label{subsec:profiles}

To simulate students who are representative in terms of
mastery levels, we define five archetypal profiles P1-P5 (Table~\ref{tab:profiles}), each of
which is one concrete SSKG. 

Each profile is specified in two coupled forms, the natural-language archetype (second column
of Table~\ref{tab:profiles}) and the SSKG itself. 
To instantiate a profile's SSKG, the mastery probability
$p_\pi(t)$ of each of the 864 CKG triples is drawn independently from the band distribution in
the third column, and the known-gap execution rate $p_{\mathrm{gap},\pi}$ is drawn once from
the profile's own band (the later-chapter band for P4 and P5).

The five archetypes divide into two groups by the structure of their mastery vectors. P1-P3
are uniform-mastery archetypes spanning near-expert to struggling. 
P4 and P5 are diagnostic.
Both sit at an intermediate overall mastery level but concentrate their gap in
complementary regions, early-chapter foundations (chapters 1-2, 214 of the 864 triples) versus
later-chapter advanced material (chapters 3-10). The chapter threshold is chosen so that the
379 items split into a balanced 140 early-only, 143 mixed, and 96 later-only partition by
their chains.

\begin{table}[!h]
\centering
\caption{Student profiles as SSKG snapshots: natural-language archetype and per-triple mastery generation.}
\label{tab:profiles}
\scriptsize
\begin{tabular}{p{0.5cm}p{2.0cm}p{4.1cm}}
\toprule
Profile & Archetype & Mastery Probability Generation \\
\midrule
P1 & near-expert & $U(0.85,0.99)$ all triples  \\
P2 & average, inconsistent & $U(0.50,0.70)$ all triples  \\
P3 & struggling & $U(0.10,0.30)$ all triples  \\
P4 & prerequisite gap & $U(0.10,0.25)$ early, $U(0.60,0.80)$ later  \\
P5 & strong basics, weak advanced & $U(0.80,0.95)$ early, $U(0.15,0.35)$ later  \\
\bottomrule
\end{tabular}
\end{table}

%
%
\section{Implementation and Theoretical Foundations of the SSKG-Based Method}
\label{sec:simulation}

The SSKG-based method solves the \textbf{Problem of Faithful Simulation} 
by separating \emph{deciding} from \emph{narrating}. Given an item's required KG triple chain 
$\mathrm{chain}(q)$ and a profile's mastery vector $p_\pi$, an analytic sampling procedure 
first decides the outcome, which chain steps the profile succeeds or fails at, and, on failure, 
which specific wrong choice results, entirely without invoking an LLM. Only afterward is an 
LLM invoked to narrate a first-person rationale consistent with the outcome. 

\subsection{Two-stage architecture}

The method consists of the following two separate stages:
\begin{align}
\mathrm{Stage\_1}: y(q,\pi) &= \mathcal{S}_1\big(\mathrm{chain}(q),\, \pi,\, \sigma\big), \label{eq:stage1}\\
\mathrm{Stage\_2}: r(q,\pi) &= \mathcal{S}_2\big(y(q,\pi),\, \tau(q,\pi,\sigma)\big), \label{eq:stage2}
\end{align}
where $\sigma$ is a recorded seed and $\tau(q,\pi,\sigma)$ is the $\mathrm{Stage\_1}$ trace 
(which chain steps succeeded, which failed and how). $\mathcal{S}_1$ is deterministic 
given $\sigma$ and requires no LLM call. It samples whether each triple in the chain is answered 
correctly and combines these draws into an overall outcome and, on failure, an answer choice. 
$\mathcal{S}_2$ invokes an LLM, supplying it the serialized SSKG 
scoped to $q$'s chain, the trace $\tau(q,\pi,\sigma)$, and the outcome $y(q,\pi)$. The LLM's 
task is to narrate a first-person rationale consistent with that trace. 

\subsection{Theoretical foundations}

The method rests on two long-standing ideas from student modeling.
The first idea, from Corbett and Anderson's Bayesian Knowledge Tracing~\cite{corbett1995knowledge},
is that a student can fail a step for two different reasons. The required piece of knowledge may
not be available at all (\emph{retrieval}), or it may be available and still be misapplied through
a careless error (a \emph{slip}). $\mathrm{Stage\_1}$ models
each chain step $i$ accordingly, with two Bernoulli draws,

\begin{align}
\mathrm{knows}_i &\sim \mathrm{Bernoulli}(p_i), \notag\\
\mathrm{slips}_i &\sim \mathrm{Bernoulli}(s) \ \text{if } \mathrm{knows}_i=1,
\label{eq:kt}
\end{align}

where $p_i = p_\pi(t_i)$ is the profile's mastery probability for the step's triple, and $s$ is a
single small, profile-independent slip probability which is the chance that a student who does know a
triple nevertheless misapplies it. We set $s=0.05$, about one careless error per twenty known
steps. Step $i$ succeeds iff $\mathrm{succ}_i = \mathrm{knows}_i\wedge\neg\mathrm{slips}_i$.
\comment{
The payoff of the two-draw decomposition is that every failure carries a \emph{kind}, that is,
a recorded cause: a retrieval failure ($\mathrm{knows}_i=0$, the knowledge was missing) or an
execution failure ($\mathrm{knows}_i=1\wedge\mathrm{slips}_i=1$, the knowledge was there but
slipped in use). A single-draw model (\texttt{arm-a} below) can record only an
\emph{undifferentiated miss}: the step failed, with no indication of why. The kind is what
$\mathrm{Stage\_2}$ narrates and what the diagnosability property (Prop3) needs.
}

The second idea comes from cognitive diagnosis, a psychometric tradition that explains a
student's item responses by which discrete skills (\emph{attributes}) the student has mastered,
rather than by a single overall ability score. Its baseline model is DINA, the Deterministic
Input, Noisy ``And'' gate model~\cite{junker2001cognitive}. Each item requires a specific set
of attributes, and the student answers correctly only if \emph{every} required attribute is
mastered, an ``and'' gate with slip and guess noise on top. The rule is conjunctive rather than
compensatory. One missing attribute sinks the item, and strength elsewhere cannot make up for it.
$\mathrm{Stage\_1}$ adopts exactly this rule, with CKG triples playing the role of attributes
and $\mathrm{chain}(q)$ the role of the item's required set,

\begin{equation}
\mathrm{correct}(q,\pi) = \prod_{i\,:\,t_i\in C(q)} \mathrm{succ}_i,
\label{eq:conjunctive}
\end{equation}

where $C(q)\subseteq\mathrm{chain}(q)$ is the \emph{active set} of steps that count toward
correctness. This product is the construction of the method-agnostic
$\mathrm{correct}(q,\pi)$. $\mathrm{Stage\_1}$ outputs
the correct choice $y(q,\pi)=y^*(q)$ directly on success, and it selects a choice from
$\mathrm{choices}(q)\setminus\{y^*(q)\}$ on failure. 

Plain DINA treats every required attribute as equally necessary, $C(q)=\mathrm{chain}(q)$.
That is too blunt for a curriculum graph whose triples differ in type. Forgetting a vocabulary
definition should not sink an item the way failing a procedural step does. G-DINA, de la
Torre's generalized DINA framework~\cite{delatorre2011gdina}, relaxes exactly this assumption
by letting different attributes contribute to correctness with different weights instead of
through an all-or-nothing gate. We modify
Eq.~\eqref{eq:conjunctive} with a type-weighted rule. Each definitional
triple has its mastery compressed toward the top,

\begin{equation}
p_{\mathrm{def}} = 1 - \kappa_{\mathrm{def}}(1-p_i), \qquad \kappa_{\mathrm{def}}\in(0,1),
\label{eq:pdef}
\end{equation}

and is dropped from the active set, $C(q) = \mathrm{chain}(q)\setminus\{\text{definitional }t_i\}$,
so a definitional failure can no longer change the outcome and surfaces only in the narration,
while procedural and relational triples retain their full discriminating power and stay in
$C(q)$. In G-DINA terms, definitional triples are demoted from necessary attributes to
low-discrimination ones.

\subsection{The four-arm ablation ladder}

Table~\ref{tab:arms} summarizes the four ablation arms. Each arm adds 
one new mechanism to the previous arm.

\begin{table}[!h]
\centering
\caption{The four-arm ablation ladder. Each arm adds one mechanism on top of the previous.}
\label{tab:arms}
\footnotesize
\begin{tabular}{lp{7cm}}
\toprule
Arm & Adds \\
\midrule
a & Single draw $\mathrm{ok}_i\sim\mathrm{Bernoulli}(p_i)$ per triple, raw taxonomy-blind 
$p_i$, guess floor 0.25 on failure. \\
b & Retrieval/execution decomposition: $\mathrm{knows}_i\sim\mathrm{Bernoulli}(p_i)$, 
then $\mathrm{slips}_i\sim\mathrm{Bernoulli}(s{=}0.05)$; still taxonomy-blind $p_i$. \\
c & Taxonomy awareness: definitional $p_i\to p_{\mathrm{def}}$ per Eq.~\eqref{eq:pdef} 
($\kappa_{\mathrm{def}}{=}0.3$); active set $C(q)$ in Eq.~\eqref{eq:conjunctive} drops to 
procedural/relational/gap steps (definitional failures are narration-only). \\
d & Distractor routing: an execution failure at a mapped triple selects that triple's specific 
wrong-answer choice instead of a uniform-random one. \\
\bottomrule
\end{tabular}
\end{table}

For each $(\text{item},\text{profile})$ pair, we generate random draws for every 
step in the solution chain using a single recorded seed $\sigma$. One draw determines 
whether the student knows the required triple, and another determines whether the 
student makes a slip. The same random draws are reused across all four arms using 
common random numbers ($\sigma$). This allows us to compare the arms on the same item 
and profile under identical random conditions.

\subsection{Illustrative Example, Continued}

Consider the illustrative example item from Section~\ref{sec:example} under profile P3, 
which has a mean per-triple mastery of 0.201. Suppose the seeded draw shows 
that the student fails triple T150, a procedural triple involving a step in unit conversion.

In \texttt{arm-a} and \texttt{arm-b}, any failure in the solution 
chain makes the answer incorrect, without distinguishing the type of 
failure. In \texttt{arm-c}, the failure still makes the answer incorrect 
because T150 is procedural rather than definitional. However, the model 
now records the failure as an \emph{execution} error.

In \texttt{arm-d}, the failure is also linked to a specific distractor. 
For this item, an execution error at T149 or T150 leads to choice B, $\frac{25}{4}m=5$. 
This represents a realistic unit-conversion mistake. Dividing miles by dollars 
instead of multiplying by a unit fraction that cancels gallons. The model 
therefore selects choice B rather than choosing randomly among the incorrect answers.

The LLM is then asked to generate a rationale that reflects this specific 
mistake and carries it through to choice B. For example, it should describe 
the unit-conversion error rather than simply saying, "I made an arithmetic error."
By contrast, a failure at T19, which is a definitional triple, would not make 
the answer incorrect in \texttt{arm-c} or \texttt{arm-d}. Instead, it would 
appear only as uncertainty or hesitation in the student's explanation.

%
%
\section{Evaluation}
\label{sec:evaluation}

We evaluate the proposed SSKG-based method against seven published
baselines and one scatter control. Our goal is the assess how well the proposed 
method performs against the baselines in terms of the three properties 
of the \textbf{Problem of Faithful Simulation} in Section ~\ref{sec:problem}. 
We also run an ablation study to analyze the mechanism that 
contributes gains. 

\subsection{Baselines for Comparison}
\label{sec:baselines}

Table~\ref{tab:baselines} lists and describes the baselines in our evaluation. The \texttt{Method}
column indicates which references the baseline is presented.

\begin{table}[!h]
\centering
\caption{Baselines for Comparison}
\label{tab:baselines}
\scriptsize
\begin{tabular}{p{0.6cm}p{3.9cm}p{2.9cm}}
\toprule
ID & Method & What controls ability \\
\midrule
B1 & Vanilla LLM, no student description & nothing \\
B2 & Persona prompt~\cite{luwang2024generative,srivatsa2025can} & the prompt \\
B3 & Level prompt, 1 to 5~\cite{benedetto2024using} & the prompt \\
B4a & Skill-list prompt, student voice~\cite{luwang2024generative} & the prompt \\
B4b & Skill-list prompt, teacher predicts student~\cite{luwang2024generative} & the prompt \\
B5 & Persona plus chain-of-thought~\cite{wei2022chain} & the prompt \\
B6 & Persona plus two worked examples & the prompt \\
B7 & Scatter control, same error count as ours & nothing \\
\textbf{Ours} & SSKG-based simulation (\texttt{arm-d}) & per-triple mastery vector \\
\bottomrule
\end{tabular}
\end{table}

B1 asks LLM to assess the SAT questions without students' descriptions. B2
adds students' profile in the prompt. 
B3 uses Benedetto et al.'s published prompt in~\cite{benedetto2024using}.
Because B3's prompt cannot express the structured gaps of profiles P4 and P5, 
the baseline B3 only runs on P1 to P3. 
B4a and B4b describe the student as a list of the five College Board skills in the SAT
dataset, each marked \texttt{mastered}, \texttt{shaky}, or \texttt{not learned}~\cite{luwang2024generative}.
B5 applies chain of thought in prompt and B6 adds two worked examples. Since we
have no real student records, we reserve two items for simulated responses as the examples. 
Finally, B7 is a scatter control. In particular, for each profile, B7 makes the same number of wrong answers as our
proposed method, but places them on random items with random wrong choices. B7 tests whether a metric
rewards the \emph{placement} of errors or only their \emph{count}. 

\subsection{Dataset and Model}
\label{sec:setup}

We use the same set of 379 SAT algebra multiple-choice questions 
retrieved from the official College Board educator question bank~\cite{collegeboard2026sat}
and the five student profiles
described in Section~\ref{subsec:profiles} (Table~\ref{tab:profiles}). Since we need to reserve
two items as B6's worked examples, the comparison uses the other 377 items. 
We run B1 and B2 with three different models from three different vendors, 
\emph{Gemini 3.1 Flash-lite}, \emph{Claude Haiku 4.5}, and \emph{GPT-5.4-mini}. Since the results do not 
depend on the vendor, we use Gemini 3.1 Flash-lite only as the LLM for other comparison 
studies. 

\subsection{Metrics}
\label{sec:metrics}

We use four metrics, M1 to M4, as described below. M1 and M2 test the property, Prop1, of the
\textbf{Problem of Faithful Simulator} as presented 
in Section~\ref{sec:problem}. M3 tests the property, Prop2. M4 tests the property, Prop3. 

\textbf{M1: accuracy by profile (Prop1).} Given a question set $Q$ and a student profile $\pi$,
let $y(q,\pi)$ represent the choice made by a simulation method on a question $q\in Q$ for the profile $\pi$.
Let $y^*(q)$ be the correct choice of the question $q$, then, the accuracy of
the method on the profile $\pi$, $\mathrm{Accuracy}(\pi)$, is defined as:

\begin{equation}
\mathrm{Accuracy}(\pi) = |Q|^{-1}\textstyle\sum_{q\in Q} \mathds{1}[y(q,\pi) = y^*(q)].
\label{eq:m1}
\end{equation}
We expect that faithful simulator would produce ordered accuracies
spread across the profiles with different mastery levels.

\textbf{M2: mastery ordering (Prop1).} We borrow 
Benedetto et al.'s metric in ~\cite{benedetto2024using} to measure the order of the accuracies. 
Let $L=(a_{P1},a_{P2},a_{P3})$ be a
method's accuracies on the three profiles, P1, P2, and P3. Let $I=(1,0.5,0)$ be an ideal, evenly spaced
curve. Then,
\begin{align}
M &= \rho(L, I) - P, \label{eq:m2}\\
P &= \textstyle\sum_{i} \sqrt{|\Delta a_i|}\cdot\mathds{1}\big[\mathrm{sign}(\Delta a_i)\neq\mathrm{sign}(\Delta I_i)\big]. \notag
\end{align}
Here $\rho$ is the Pearson correlation, and 
$\Delta a_i$ and $\Delta I_i$ are the difference of accuracies between adjacent profiles.
$P$ penalizes the correlation if adjacent
profiles moves in different way. 

\textbf{M3: gap localization (Prop2).}
This metric assesses whether a structured-gap profile, such as P4 or P5, makes more errors on 
its weak skills than on its strong skills. In the College Board question bank~\cite{collegeboard2026sat},
each SAT question is labeled by a skill. There are five labels for the 
SAT algebra questions as shown in Table~\ref{tab:skillstatus}, where each skill is marked as 
\texttt{mastered}, \texttt{shaky}, or \texttt{not learned}. 

\begin{table}[!h]
\centering
\caption{Weak Skills (W) and Strong Skills (S)}
\label{tab:skillstatus}
\scriptsize
\begin{tabular}{lll}
\toprule
College Board skill & Profile P4 & Profile P5 \\
\midrule
Linear equations in one variable & \texttt{not learned} (W) & \texttt{mastered} (S) \\
Linear equations in two variables & \texttt{shaky} (S) & \texttt{not learned} (W) \\
Linear functions & \texttt{shaky} (S) & \texttt{not learned} (W) \\
Systems of two linear equations & \texttt{shaky} (S) & \texttt{not learned} (W) \\
Linear inequalities & \texttt{shaky} (S) & \texttt{not learned} (W)  \\
\bottomrule
\end{tabular}
\end{table}

As the profile's design in Table~\ref{tab:profiles}, P4 is weak on early materials, so its
first skill, \emph{Linear equation in one variable}, is \texttt{not learned}.
P5 is strong on early materials and weak on later materials, so it has the opposite
pattern. A \emph{weak skill} is a skill marked as \texttt{not learned}.
A \emph{strong skill} is a skill marked as \texttt{mastered}. P4 has no \texttt{mastered} skill,
so its strong skills are the ones marked as \texttt{shaky}, its least-weak skills.

Given a simulation method and a student profile, let $Q_W$ be the set of questions labeled
with a weak skill, and let $Q_S$ be the set of questions labeled with a strong skill.
We compute the error rate of a set of questions as the percentage of 
wrong answers by the method for the profile. For $X\in\{W,S\}$,
\begin{equation}
\mathrm{ErrorRate}(Q_X) = \frac{1}{|Q_X|}\sum_{q\in Q_X} \mathds{1}[y(q,\pi) \neq y^*(q)].
\label{eq:errrate}
\end{equation}

Then, the gap metric, M3,  is computed as:
\begin{equation}
\mathrm{gap} = \mathrm{ErrorRate}(Q_W) - \mathrm{ErrorRate}(Q_S).
\label{eq:m3}
\end{equation}

We expect a positive gap, which means that the method makes more
errors when the profile is on weak side. A near-zero gap means
errors fall evenly regardless of the profile's skill sets.

\textbf{M4: mastery-error correlation (Prop3).} 
This metric assesses whether low mastery of a specific KG triple leads to
more errors on the questions that require the triple. If it does, a wrong answer
can be traced back to weak knowledge. 

We compute the metric, M4, in four steps.
\begin{enumerate}
\item For each triple $t$, collect the questions that need it,
$Q_t = \{\,q : t \in \mathrm{chain}(q)\,\}$.
\item For each profile $\pi$, take the triple's mastery $p_\pi(t)$ from the SSKG. 
\item Under the same profile, compute the error rate on those items,
$e_\pi(t) = \mathrm{ErrorRate}(Q_t)$, using Eq.~\eqref{eq:errrate}.
\item Keep the pairs $(t,\pi)$ with $|Q_t| \ge 3$, since an error rate over fewer items is
too noisy. Pool the kept pairs over all five profiles.
\end{enumerate}
M4 is the Spearman rank correlation over the kept pairs,
\begin{equation}
\mathrm{M4} = \rho_{\mathrm{S}}\Big(\big\{p_\pi(t)\big\},\ \big\{e_\pi(t)\big\}\Big).
\label{eq:m4}
\end{equation}
The expected direction is negative. Lower mastery should go with a higher error rate. 
As a finer check, we also
pair each (item, profile) with the lowest mastery in its chain, called the
\emph{bottleneck}, and correlate it with whether the item was answered correctly.
M4 applies only to our
method, since the baselines have no per-triple mastery. 

\subsection{Evaluation Results}
\label{sec:comparison}

\subsubsection{Monotonicity (Prop1)}

Table~\ref{tab:accuracy} shows the accuracies measured by M1. 
The seven baselines score between
86.5\% and 100\% on every profile. B1, B2, B4a, B5, and B6 stay between 96.6\% and 100\%,
even under P3, whose prompt describes a student with low mastery of most algebra concepts.
B5 reaches 100\% on every profile. 

\begin{table}[!h]
\centering
\caption{Accuracy by method and profile (Eq.~\ref{eq:m1})}
\label{tab:accuracy}
\scriptsize
\begin{tabular}{lrrrrr}
\toprule
Method & P1 & P2 & P3 & P4 & P5 \\
\midrule
B1 & 1.000 & 1.000 & 1.000 & 1.000 & 1.000 \\
B2 & 1.000 & 0.997 & 0.968 & 1.000 & 1.000 \\
B3 & 0.987 & 0.984 & 0.897 & n/a & n/a \\
B4a & 1.000 & 0.997 & 0.968 & 0.997 & 0.992 \\
B4b & 1.000 & 0.894 & 0.865 & 0.912 & 0.976 \\
B5 & 1.000 & 1.000 & 1.000 & 1.000 & 1.000 \\
B6 & 1.000 & 0.997 & 0.966 & 0.995 & 1.000 \\
\midrule
B7 (scatter) & 0.859 & 0.578 & 0.446 & 0.475 & 0.602 \\
\textbf{Ours} & \textbf{0.852} & \textbf{0.578} & \textbf{0.441} & \textbf{0.485} & \textbf{0.599} \\
\bottomrule
\end{tabular}
\end{table}

The ceiling effect is not specific to one vendor. For the baseline B1, all LLMs scores 100\%.
For B2, Gemini scores 96.8\% to 100\% across the five profiles. Claude~Haiku~4.5 and GPT-5.4-mini 
are flatter still, at 99.2\% to 99.7\% on every profile.

B4b spreads the most from 86.5\% to 100\%. This agrees with Lu and Wang's
finding that asking an LLM to \emph{predict} a student works better than asking it to
\emph{be} the student~\cite{luwang2024generative}. Still, B4b stays within 14 percentage
points of the ceiling.

Our method spans 44.1\% to 85.2\%, a range of 41 percentage points. On the mastery-ordered
profiles, it falls from 85.2\% (P1) to 57.8\% (P2) to 44.1\% (P3). 
Our method also ranks P5 above P4, although P5 has the lower mean
mastery. P4 is weak on the early chapters, and most item chains pass through early
material. 

B7 matches our accuracy closely, as designed. So M1 alone cannot tell a faithful simulator
from random errors at the right rate. We will see later that the metric M3 
will distinguish our method from the scatter control B7.

Our method scores highest on M2, at 0.982 (Table~\ref{tab:m2}). B1 and B5 score 0, since
their accuracy does not change across profiles. M2 uses correlation, which looks at the
direction of change and ignores its size. So B2, B4a, and B6 score about 0.90 even though
their accuracy moves by only a few points. B7 scores 0.979, because it is also strictly
ordered. With only three profiles, M2 cannot separate a wide spread from a narrow one. The
spread in M1 is the stronger evidence. Our method is the only one with both the right order
and wide gaps between profiles.

\begin{table}[!h]
\centering
\caption{M2, mastery ordering (Eq.~\ref{eq:m2}), with bootstrap 95\% CI.}
\label{tab:m2}
\scriptsize
\begin{tabular}{lrl}
\toprule
Method & $M$ & 95\% CI \\
\midrule
B1 & 0.000 & [0.000, 0.000] \\
B2 & 0.901 & [0.866, 0.990] \\
B3 & 0.879 & [0.659, 0.950] \\
B4a & 0.901 & [0.866, 0.961] \\
B4b & 0.950 & [0.700, 0.999] \\
B5 & 0.000 & [0.000, 0.000] \\
B6 & 0.899 & [0.866, 0.966] \\
B7 (scatter) & 0.979 & [0.929, 0.999] \\
\textbf{Ours} & \textbf{0.982} & \textbf{[0.944, 1.000]} \\
\bottomrule
\end{tabular}
\end{table}

\subsubsection{Skill specificity (Prop2)}
\label{sec:res-prop2}

\begin{table}[!h]
\centering
\caption{M3, gap localization (Eq.~\ref{eq:m3}). A positive gap is the expected direction.}
\label{tab:m3}
\scriptsize
\begin{tabular}{lrr}
\toprule
Method & P4 gap & P5 gap \\
\midrule
B1 & +0.000 & +0.000 \\
B2 & +0.000 & +0.000 \\
B3 & n/a & n/a \\
B4a & $-$0.003 & +0.010 \\
B4b & +0.010 & +0.030 \\
B5 & +0.000 & +0.000 \\
B6 & $-$0.007 & +0.000 \\
B7 (scatter) & +0.062 & +0.001 \\
\textbf{Ours} & \textbf{+0.217} & \textbf{+0.079} \\
\bottomrule
\end{tabular}
\end{table}

M3 separates our method from every other method, including B7 (Table~\ref{tab:m3}). Our gap
is +0.217 for P4 and +0.079 for P5. B7 has the same number of errors but places them at
random, so its gap is close to zero. Our P4 gap is more than three times B7's and more than
seven times the largest prompting-baseline gap (B4b's +0.030). B4a and B4b receive a skill
list, but the LLM still answers almost every item correctly, so their gaps stay near zero.

Table~\ref{tab:skillspec} breaks M3 down by chain footprint, over all 379 items. B2 is at
100\% in every cell. For P4, our method is weak on early and mixed items and 35.7 points
stronger on later-only items, as expected. P5 is strongest on early items (70.0\%), as
designed. Its weakest group is mixed (48.3\%), not later-only (62.5\%). Mixed chains are
longer, and the item is correct only if every required step succeeds. So each extra step is
one more chance to fail, which adds a length penalty to P5's weak later steps.

\begin{table}[!h]
\centering
\caption{M3 breakdown. Accuracy by chain footprint (140 early-only, 143 mixed, 96
later-only items).}
\label{tab:skillspec}
\scriptsize
\begin{tabular}{llrrr}
\toprule
Profile & Method & Early & Mixed & Later \\
\midrule
P4 (weak early) & B2 & 100.0\% & 100.0\% & 100.0\% \\
 & \textbf{Ours} & 39.3\% & 39.9\% & \textbf{75.0\%} \\
P5 (weak later) & B2 & 100.0\% & 100.0\% & 100.0\% \\
 & \textbf{Ours} & \textbf{70.0\%} & 48.3\% & 62.5\% \\
\bottomrule
\end{tabular}
\end{table}

\subsubsection{Diagnosability (Prop3)}

For M4, we pooled the 525 (triple, profile) pairs in which the triple appears in at least
three item chains. A triple's mastery correlates with the error rate on those items at
Spearman $\rho=-0.459$ ($p\approx10^{-28}$). Lower mastery of a specific triple goes with
more errors on the items that need it. The bottleneck check over all 1,895 (item, profile)
pairs gives $\rho=0.318$ ($p\approx10^{-45}$). It is weaker, as expected, since one lowest
value cannot summarize a whole chain. Across all 1,895 of our records, the answer stated
in the rationale also matches the answer decided by Stage~1 in every case. So each failed-triple
label is carried through to the written response.

\subsection{Ablation Study}
\label{sec:ablation}

The results above show that our method meets Prop1, Prop2, and Prop3. 
The ablation study asks which of its mechanisms produce these results. 
We use the four arms of Section~\ref{sec:simulation} (Table~\ref{tab:arms}). 
Each arm adds one mechanism to the previous one. All arms use the same 
random seeds on the same 1,895 (item, profile) pairs,
so differences between arms come only from the added mechanism.

\begin{figure}[!h]
\centering
\includegraphics[width=\columnwidth]{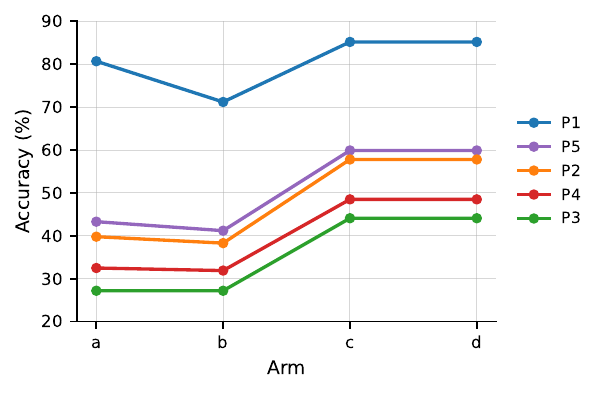}
\caption{Accuracy by profile across the four arms. The b$\to$c step makes the largest change.
\texttt{arm-c} and \texttt{arm-d} are identical in accuracy.}
\label{fig:ladder}
\end{figure}

\begin{table}[!h]
\centering
\caption{Change in accuracy at each step of the ablation.}
\label{tab:attribution}
\scriptsize
\begin{tabular}{lll}
\toprule
Step & Change & Main effect \\
\midrule
a$\to$b & $-9.5$ to $0.0$ pp & mostly P1 \\
b$\to$c & $+14.0$ to $+19.5$ pp & larger for lower mastery \\
c$\to$d & $0.0$ pp & diagnosis only \\
\bottomrule
\end{tabular}
\end{table}

Even \texttt{arm-a} spreads the five profiles over a 30 to 55 point range
(Figure~\ref{fig:ladder}). Sampling over the SSKG alone escapes the ceiling that traps every
prompting baseline. The three steps then shape that spread (Table~\ref{tab:attribution}).

\textbf{a$\to$b, execution slips.} Accuracy drops by 9.5 points for P1 and by at most 2.1
points for any other profile. All 36 items that P1 newly fails are execution failures. A
near-expert who never slips is not realistic, and this step moves P1 off that ceiling.

\textbf{b$\to$c, taxonomy awareness.} This step makes the largest change. Accuracy rises by
14.0 to 19.5 points across profiles, as failed definitions no longer sink whole items. The
rise is larger for lower-mastery profiles. P2 to P5 gain 16.6 to 19.5 points, and P1 gains
14.0. Definitional triples are common (461 of 864), and weaker profiles fail them more
often.

\textbf{c$\to$d, distractor routing.} Accuracy does not change for any profile, as
designed. In 19 of 1,895 records, an execution failure goes to its mapped wrong choice, and
the rationale describes that specific error.

\section{Discussion}
\label{sec:discussion}

Our study shows that 
\emph{prompting cannot hold a capable LLM below its own ability.} 
Across seven baselines and three different LLMs, performance on the 
SAT algebra questions consistently remained within 14\%
points of the ceiling for every profile. This finding matches
prior reports~\cite{benedetto2024using,srivatsa2025can}. In out method, sampling the SSKG
decides the answer. That is why the accuracy follows the profile.

For evaluating performance, \emph{accuracy alone is insufficient.} The baseline B7 
matches our error rate and looks almost the
same on the metrics M1 and M2. Only the metric M3, 
which asks \emph{where} errors fall, separates our SSKG-based method from B7. Future
work on student simulation should report a placement metric next to accuracy.

\emph{The proposed theoretical properties are largely supported by the experimental results}. 
For our  SSKG-based method, Prop1 is met clearly. M1 shows wide, ordered steps from P1 to P3, 
and M2 confirms the order. Prop2 is met for both structured-gap profiles under M3. Prop3 holds
at the decision level, since M4 shows that triple mastery predicts errors. 

\emph{Each mechanism contributes differently to the gains of the proposed SSKG-based method.}
Execution slips reduce the unrealistically high accuracy of near-expert profiles. 
Taxonomy awareness makes the largest gain, especially for weaker
profiles. Distractor routing leaves accuracy unchanged and makes some wrong 
answers more diagnostic.

%
%
\section{Limitations}
\label{sec:limitations}

This work has several limitations. The first two concern the method. The rest concerns the
evaluation in Section~\ref{sec:evaluation}.

\textbf{1. Manual effort.} Building the CKG, the triple chains, and the distractor maps takes
substantial human effort, even with LLM help. More automated construction would make the
method easier to apply to new curricula.

\textbf{2. No real student data.} The evaluation shows that our accuracy falls in the right
order across profiles and that errors land on weak skills. It does not show how closely the
simulated accuracy matches real students with the same mastery. The five profiles are also
designed by hand and are not fitted to real students. Validating the simulation against real
student responses is an important next step.

\textbf{3. Simplified baselines.} Our skill-list baselines (B4a, B4b) use a plain table of
skills marked \texttt{mastered}, \texttt{shaky}, or \texttt{not learned}. Lu and Wang's original prompts also include
worked examples for each knowledge component~\cite{luwang2024generative}. Our B6 examples
come from our own simulator, since we have no real student work. Readers comparing directly
with published numbers should keep these choices in mind.

\textbf{4. One LLM in the main comparison.} All baselines in the main comparison use
Gemini~3.1~Flash-Lite. Using different and advanced LLMs is a future plan. 

%
%
\section{Conclusion}
\label{sec:conclusion}

Our results show that direct natural-language prompting does not reliably 
simulate students at different 
mastery levels. We address this limitation by moving the mastery model outside the 
prompt and into a Stochastic Student Knowledge Graph (SSKG). 
The simulation samples mastery over a curriculum-derived chain of 
required triples to determine the outcome before the LLM generates a 
response. Compared to seven published or standard baselines on the same test items, 
profiles, and LLM, this SSKG-based method restores a monotone mastery gradient, produces 
skill-specific error patterns, and assigns diagnostic labels to every 
synthetic response by construction. The computational cost also scales mainly 
with the amount of generated narration rather than the number of decisions.

Several directions remain for future work. First, this study's prompting baselines could not
include a model-level baseline such as PS$^2$~\cite{liu2026ps2}, which needs a fine-tuned weak
model and GPU access, or an architectural baseline such as BEAGLE~\cite{wang2026beagle}, whose
domain (longitudinal, open-ended coding trajectories) is disjoint from single-shot multiple-choice
items. A compute budget or a shared benchmark that accommodates both would let the comparison
include model and architectural controls. 
Second, simulated performance and item difficulty should be validated against
external Item Response Theory statistics and real student data, rather
than relying only on the internal metrics used here. Third,
the corpus and simulation framework can be extended to support downstream
knowledge-tracing models trained on the by-construction diagnostic labels.
These steps would connect the present work to a broader research program
on stochastic knowledge graphs and support the development of a
benchmark for evaluating faithful student simulation.

\bibliographystyle{IEEEtran}


\end{document}